\documentclass[letterpaper, 10 pt, conference]{ieeeconf}

\usepackage{amssymb,amsmath,color,amsfonts}
\usepackage{graphicx}
\usepackage{caption}
\usepackage{cite}
\IEEEoverridecommandlockouts                              
\newcommand*{\QEDB}{\hfill\ensuremath{\square}}%

\newtheorem{assumption}{Assumption}
\newtheorem{definition}{Definition}

\newtheorem{lemma}{Lemma}
\newtheorem{theorem}{Theorem}

\newtheorem{proposition}{Proposition}
\newtheorem{remark}{Remark}

\begin{document}

\title{\bf  Data-Assisted Vision-Based Hybrid Control for Robust Stabilization with Obstacle Avoidance via Learning of Perception Maps }

\author{Alejandro Murillo-González\thanks{Alejandro Murillo-González is with the Department of Mathematical Sciences at Universidad EAFIT, Colombia. Email \{amurillog@eafit.edu.co\}}, Jorge I. Poveda, \thanks{Jorge I. Poveda is with the Department of Electrical, Computer, and Energy Engineering, University of Colorado, Boulder, CO, 80309. Email:\{jorge.poveda@colorado.edu\}.}}

\maketitle
\begin{abstract}
We study the problem of target stabilization with robust obstacle avoidance in robots and vehicles that have access only to vision-based sensors for the purpose of real-time localization. This problem is particularly challenging due to the topological obstructions induced by the obstacle, which preclude the existence of smooth feedback controllers able to achieve simultaneous stabilization and robust obstacle avoidance. To overcome this issue, we develop a vision-based hybrid controller that switches between two different feedback laws depending on the current position of the vehicle using a hysteresis mechanism and a data-assisted supervisor. The main innovation of the paper is the incorporation of suitable \emph{perception maps} into the hybrid controller. These maps can be learned from data obtained from cameras in the vehicles and trained via convolutional neural networks (CNN). Under suitable assumptions on this perception map, we establish theoretical guarantees for the trajectories of the vehicle in terms of convergence and obstacle avoidance. Moreover, the proposed vision-based hybrid controller is numerically tested under different scenarios, including noisy data, sensors with failures, and cameras with occlusions.
\end{abstract}
\section{Introduction}
During recent years, there has been an increasing number of works on systems that integrate high-dimensional inputs, such as images, into feedback control loops. For example, several successful end-to-end approaches have employed reinforcement learning (RL), including \cite{finn16}, where the state-space construction is automated by learning a state representation directly from camera images.  Also, in \cite{mnih15} the authors introduced deep Q-networks into a control command, achieving approximate human-level performance. Other works have used deep generative models to synthesize controllers with inputs coming from an embedding space of high-dimensional data, which does not necessarily correspond to an interpretable space (e.g., joint coordinates of the robot). Some examples in this direction include \cite{wahlstrom15, watter15, banijamali18, lambert18, hafner19}. In other works, such as \cite{banijamali18, dean19} and \cite{petrik2020learning}, the authors integrated state predictions via robust control tools to handle approximation errors. 

While significant progress has been made in different communities during the last years, most of the results in the literature have focused on applications where the feedback control law $\alpha(\cdot)$ leads to closed-loop systems of the form
\begin{equation}\label{smoothfeedback}
    \dot{x}=f(x+e_1,u)+e_2,~~u=\alpha(x+e_3)+e_4,
\end{equation}
where $f$ and $\alpha$ are continuous functions, and the signals $e_i$, for $i\in\{1,2,3,4\}$, model measurement noise, implementation errors, or approximation inaccuracies induced by learning mechanisms such as linear parametric approximations, neural networks, multi-time scale techniques, etc. For these perturbed dynamical systems, stability and robustness results are well established, and they can be characterized via practical or input-to-state stability results \cite{khalil,sontag_robustness_noise}.

On the other hand, many robust control problems cannot be solved via smooth dynamical systems of the form \eqref{smoothfeedback}. Typical examples include robust global stabilization problems on smooth compact manifolds \cite{Mayhew10Thesis}, global stabilization of a disconnected set \cite{sanfelice2006}, the asymptotic stabilization of vehicles with geometric constraints \cite{sontag_robustness_noise}, the robust control of switched systems \cite{Goebel:12}, and the \emph{robust stabilization of targets in obstacle avoidance problems} \cite{poveda2021}, to name just a few. In the latter problem, the objective is to robustly stabilize a target point in spaces with global obstacles, i.e., the operational space, or ``free world'', is a strict subset of $\mathbb{R}^n$. In this setting, global stabilization using smooth feedback of the form \eqref{smoothfeedback} is precluded by the fact that the domain of attraction of an asymptotically stable vector field (i.e., the operational space) must be diffeomorphic to the Euclidean space, a condition that is not satisfied under global obstacles \cite{sontag_robustness_noise}. Given that discontinuous controllers have also been shown to suffer from fundamental robustness limitations \cite{Mayhew10Thesis}, most works have focused on achieving local or almost global convergence results \cite{koditschek1990robot,durr_obstacles,ObstacleConvexPotentials,Obstacle_Zuyev,StochasicPotentialField}, which exclude from the basin of attraction a particular set of measure zero. On the other hand, the impossibility result for smooth controllers has also triggered an active line of research on the development of hybrid control techniques able to achieve robust \emph{global} stabilization and obstacle avoidance, e.g., \cite{sanfelice2006}, \cite{poveda2021}, \cite{AvoidanceBisoffi}, \cite{KellettObstacle21}, \cite{Strizic:17_CDC}. However, unlike smooth dynamical systems of the form \eqref{smoothfeedback}, establishing suitable robustness guarantees for hybrid controllers is far from trivial, which motivates current research on the integration and analysis of learning-based mechanisms into these types of systems. In particular, to the best knowledge of the authors, the systematic integration of hybrid control and data-assisted vision-based mechanisms for robust stabilization and obstacle avoidance has remained an open problem.

\textsl{Contributions:} In this work, we develop a \emph{vision-based} hybrid controller for robust and resilient obstacle avoidance in mobile robots. We show that, unlike standard smooth feedback controllers, the proposed hybrid algorithm can overcome arbitrarily small and potentially adversarial disturbances, noisy states, sensor failures, as well as camera occlusions. 
Our approach synergistically leverages robust hybrid control theory \cite{Goebel:12} and recent results in perception-based control \cite{lambert18, dean19}, which have studied the incorporation of \emph{perception maps} learned from data to predict the states and dynamics of the system. The proposed hybrid controllers are suitable for vehicles with \emph{vision} sensors, such as cameras, that have access to historical data in order to learn a suitable perception map via convolutional neural networks (CNNs). Our main results provide theoretical guarantees, as well as extensive numerical validations in different scenarios.
\section{Preliminaries}
\label{sec:preliminaries}
Given a compact set $\mathcal{A}\subset \mathbb{R}^n$ and an arbitrary vector $z \in \mathbb{R}^n$, we use $|z|_{\mathcal{A}}:= \text{min}_{s \in \mathcal{A}} ||z - s||_2$ to denote the minimum distance from $z$ to $\mathcal{A}$. We use $(x,y)$ to denote the concatenation of the vectors $x$ and $y$. A set-valued mapping $M : \mathbb{R}^p \rightrightarrows \mathbb{R}^n$ is said to be: a) outer semicontinuous (OSC) at $z$ if for each sequence $\{z_i, s_i\} \rightarrow (z, s) \in \mathbb{R}^p \times \mathbb{R}^n$ satisfying $s_i \in M(z_i)$ for all $i \in \mathbb{Z}_{\geq 0}$, we have $s \in M(z)$; b) locally bounded at $z$ if there exists an open neighborhood $N_z\subset\mathbb{R}^p$ of $z$ such that $M(N_z)$ is bounded. We use $r \mathbb{B}$ to denote a closed ball in the Euclidean space, of radius $r > 0$, and centered at the origin, and we use $\{p\}+r\mathbb{B}$ to denote the union of all the points $p_i$ that satisfy $|p-p_i|\leq r$.  Given a set $B$, we use $\bar{B}$ and $\text{bd}(B)$ to denote the closure, and the boundary, respectively, and we use $\text{int}(B)$ to denote its interior. Given a single-valued or set-valued map $f$, we use $\text{dom}(f)$ to denote its domain.

In this paper, we will use the formalism of hybrid dynamical systems \cite{Goebel:12} for the synthesis and analysis of robust vision-based control systems. Specifically, a hybrid dynamical system (HDS) with state $z \in \mathbb{R}^n$ is represented by its \emph{data} $\mathcal{H}:= \{C, F, D, G\}$, and the dynamics
\begin{subequations}\label{eq:hds_1a}
\begin{align}
    &z \in C, ~~~~ \Dot{z}\in F(z),\\
    &z \in D, ~~ z^{+} \in G(z),
\end{align}
\end{subequations}
where the set-valued mappings $F: \mathbb{R}^n\rightrightarrows \mathbb{R}^n$ and $G: \mathbb{R}^n \rightrightarrows \mathbb{R}^n$, called the flow map and the jump map, respectively, describe the evolution of the state $z$ when it belongs to the flow set $C$, and the jump set $D$, respectively.  Solutions to \eqref{eq:hds_1a} are defined on \emph{hybrid time domains}, which, under mild assumptions on the data $\mathcal{H}$, permits the use of graphical convergence notions to establish sequential compactness results for the solutions of \eqref{eq:hds_1a}, e.g., the graphical limit of a sequence of solutions is also a solution. These sequential compactness results play an important role in the robustness analysis of dynamical systems. For a precise definition of hybrid time-domains and solutions to HDS of the form \eqref{eq:hds_1a} we refer the reader to \cite[Ch. 2]{Goebel:12}. 

\vspace{0.05cm}
To guarantee suitable robustness properties, we will impose the following Basic Conditions on the data $\mathcal{H}$.
\begin{definition}\label{defwellposed}
The HDS \eqref{eq:hds_1a} is said to satisfy the Basic Conditions if: (a) the sets $C\subset\text{dom}(F)$ and $D\subset\text{dom}(G)$ are closed; (b) $F$ is convex-valued, outer-semicontinuous, and locally bounded relative to $C$; (c) $G$ is outer-semicontinuous and locally bounded relative to $D$. \QEDB
\end{definition}

Note that when $F$ is a (single-valued) continuous function, item (b) of Definition \ref{defwellposed} is automatically satisfied.

\section{The Obstacle Avoidance Problem: Robustness Limitations in Smooth Vision-Based Control}
\label{sec_obstacle}
In this paper, we are interested in the synthesis and analysis of robust feedback controllers able to autonomously steer a vehicle from any initial position $p_0\in\mathbb{R}^2$ to a final target $p_T\in\mathbb{R}^2$, by using real-time data provided by a \emph{visual sensor} as feedback. Typical examples include cameras and high-dimensional data generated by the fusion of multiple noisy sensors. To illustrate our controllers, we will consider simple velocity actuated vehicle dynamics, given by an integrator evolving on the plane, of the form
\begin{equation}\label{dynamicsvehicle}
\dot{x}=u_x,~~~~\dot{y}=u_y,~~~~~\theta=h(x,y),   
\end{equation}
where $(x,y)$ are the coordinates in the Cartesian plane, and $\theta$ corresponds to real-time data generated by $h$, which can be seen as a map that produces images as functions of the vehicle's position. The main goal is to design a feedback law $(u_x,u_y)$ such that the trajectories of the vehicle avoid an obstacle $\mathcal{N}\subset\mathbb{R}^2$ contained in a sphere of constant radius, and also converge to an arbitrarily small neighborhood of the target destination $p_T\in\mathbb{R}^2$. Such types of navigation problems have been extensively studied in the literature via different approaches, including planning and tracking algorithms \cite{BiyikFormation2008,Pappas_SourceSeeking}, triangular partitions \cite{DiscreteAbstraction}, and barrier functions \cite{AmesTAC17}, to name just a few. In contrast to these settings, in this paper, we are interested in real-time feedback-based controllers where planning and navigation are simultaneously executed, and where robustness guarantees can be provided under arbitrarily small disturbances.
\begin{remark}
Even though, for simplicity, in this paper we focus on simple velocity actuated dynamics of the form \eqref{dynamicsvehicle}, our results can be easily extended to more complex models, including nonlinear and nonholonomic dynamics, by using a multi-time scale approach, where a low-level controller (smooth, or hybrid, if needed) stabilizes the vehicle with respect to an external reference, see \cite[Sec. VI]{poveda2021}. \QEDB
\end{remark}
\subsection*{Gradient Flows, Anti-Potentials, and Perception Maps}
One of the most popular approaches for the solution of navigation problems in mobile robots is based on implementing navigation functions $\phi:\mathbb{R}^2\to\mathbb{R}$, and gradient-based feedback laws of the form
\begin{equation}\label{navigation_feedback}
u_x=k_x\frac{\partial \phi(x,y)}{\partial x},~~~~~~u_y=k_y\frac{\partial \phi(x,y)}{\partial y}.       
\end{equation}
In this setting, a continuously differentiable function $\phi$ is usually designed to have a maximizer at the desired target point, while also having minimizers at the location of the obstacles. In this way, the control law \eqref{navigation_feedback} incorporates attractive terms (to converge to the target point) and repulsive terms (to avoid the obstacles); see \cite{koditschek1990robot,durr_obstacles,ObstacleConvexPotentials,Obstacle_Zuyev,StochasicPotentialField}. Note that the closed-loop dynamics \eqref{dynamicsvehicle}-\eqref{navigation_feedback} can be written as $\dot{p}=k\nabla \phi(\theta)$, with $p=(x,y)^\top$, evolving in the set $\mathbb{R}^2\backslash\mathcal{N}$, where for simplicity we used $k_x=k_y=k\in\mathbb{R}_{>0}$.

To study controllers based on vision-based sensors, and similar to \cite{dean19}, we will assume the existence of a perception map $\ell$ that generates imperfect predictions of the state of the vehicle using the images $\theta$, namely, $\ell(\theta)=Mp+e$, where $M\in\mathbb{R}^{2\times 2}$ is a constant matrix, and $e\in\mathbb{R}^2$ is the approximation error. Using this perception map to close the loop between the camera and the vehicle, the resulting dynamics become
\begin{equation}\label{closedloop1}
\dot{p}=k\nabla \phi(\ell(\theta)),   ~~~p\in\mathbb{R}^2\backslash\mathcal{N}. 
\end{equation}
To learn the perception map $\ell$, in this paper we will use data-driven techniques that make use of a sequence of labeled training data $\mathcal{T}=\{p_i,\theta_i\}_{i=1}^N$, to be used in traditional supervised learning methods (e.g., CNNs), and which is selected to satisfy the following assumption.
\begin{assumption}\label{assumotion1}
For each compact set $K\subset\mathbb{R}^2$, and each pair $L,\varepsilon>0$, there exists a function $\ell$ learned with training data $\mathcal{T}=\{p_i,\theta_i\}_{i=1}^N$, such that $K\subset \text{int}(\mathcal{S}^L_{\varepsilon})$, where $\mathcal{S}^L_{\varepsilon} :=$
\begin{equation*}
\bigcup_{(p_d,\theta_d)\in \mathcal{T}} \left\{p\in \{p_d\}+r\mathbb{B}:|\ell(\theta_d)-Mp_d|+L|p-p_d|\leq\varepsilon\right\}.
\end{equation*}
\end{assumption}

In words, Assumption \ref{assumotion1} guarantees the existence of sufficient data to learn a suitable perception map that can cover any compact set $K$ of interest. This assumption is standard in the literature of perception-based control, e.g, \cite{dean19}. It allows establishing the following lemma, which will be instrumental for the characterization of the approximation error of the perception map learned from the data. The proof follows by a straightforward application of the triangle inequality.
\begin{lemma}\label{lemmaperceptionerrorbounded}
Let $F(p):=\ell\circ h(p)-Mp$, and suppose that Assumption \ref{assumotion1} holds and $p\mapsto F(p)$ is $L$-Lipschitz. Then, $|\ell(\theta)-Mp|\leq \varepsilon$, for all $(p,\theta)$ such that $p\in\mathcal{S}^L_{\varepsilon}$. \QEDB
\end{lemma}

\begin{figure}
\centering
\includegraphics[width=0.9\linewidth]{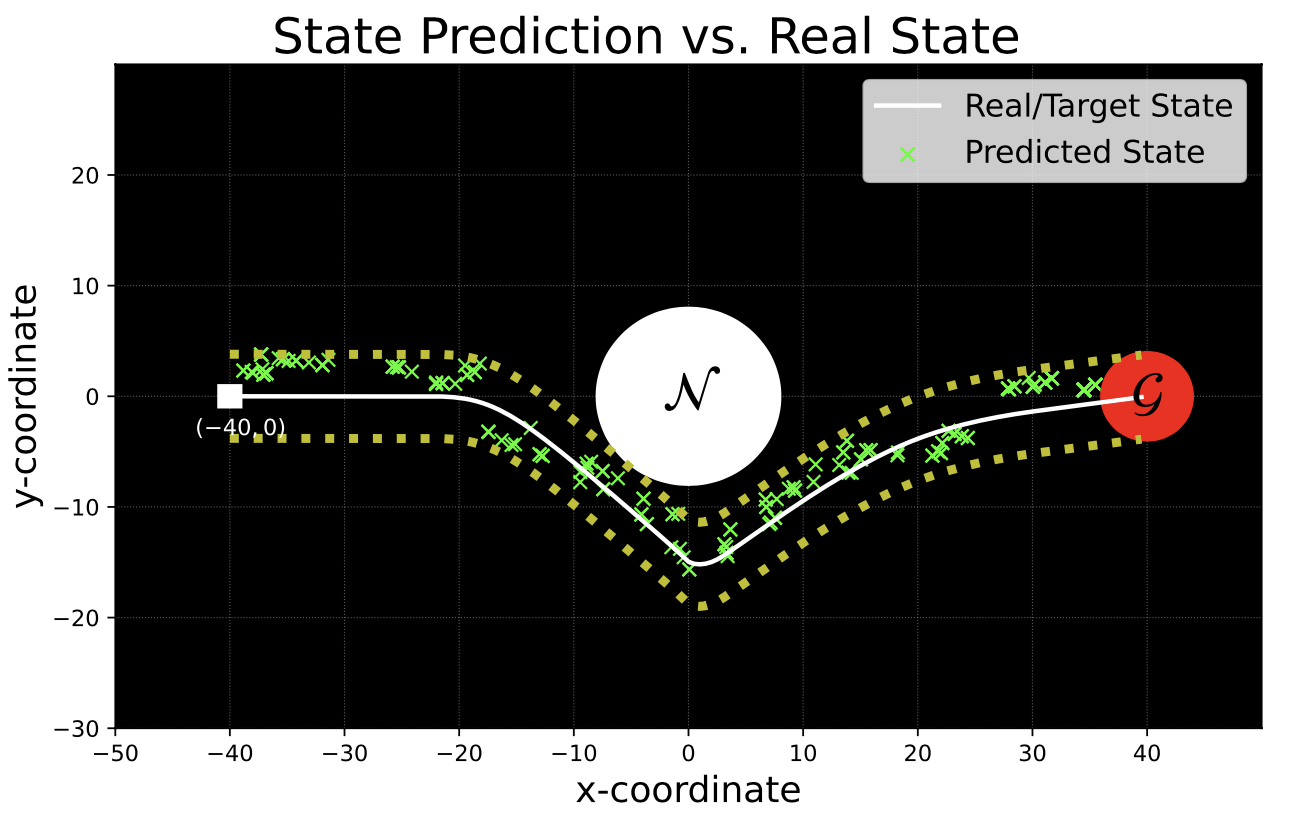}
\caption{\small{Closeness between trajectory of the robot and the predicted states by a learned perception map via convolutional neural networks (CNN).}}
\vspace{-0.4cm}
\label{fig:convergencescheme001}
\end{figure}

\vspace{0.1cm}
To illustrate Lemma \ref{lemmaperceptionerrorbounded}, Figure \ref{fig:convergencescheme001} shows a trajectory of a vehicle, as well as the predicted states by a perception map $\ell$ that satisfies $|\ell(\theta)-Mp|\leq \varepsilon$ on compact sets, with $M$ being the identity matrix. The perception map was learned by using a convolutional neural network (CNN). As observed, the predictions of the perception map remain in an $\varepsilon$-neighborhood of the actual trajectory. Therefore, throughout the rest of this document, we will take $M=I$.  

\begin{figure}
    \centering
    \vspace{0.25cm}
    \includegraphics[width=0.9\linewidth]{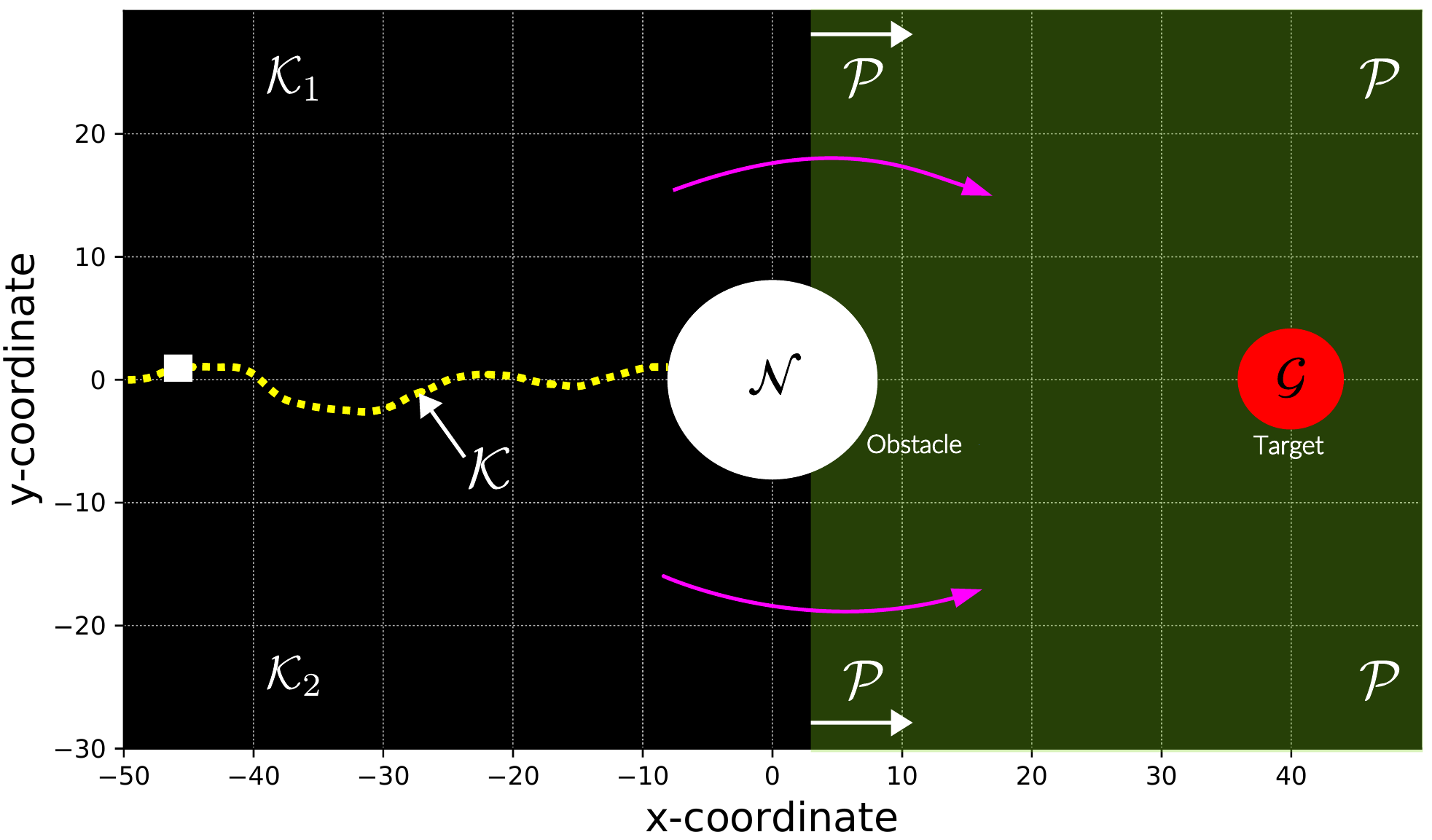}
    \caption{\small{An obstacle avoidance problem with target $\mathcal{G}$ and obstacle $\mathcal{N}$, and the ``sensitive'' set $\mathcal{K}$.}}
    \label{fig:my_label}
\end{figure}

Note that if the learned perception map $\ell$ satisfies the conditions of Lemma \ref{lemmaperceptionerrorbounded}, then the closed-loop system  \eqref{closedloop1} behaves as the following perturbed dynamical system
\begin{equation}\label{perturbed_closed}
\dot{p}=k\nabla \phi(p+e),~~~~|e|\leq \varepsilon,~~~~\forall~p\in\mathcal{S}^L_{\varepsilon}\backslash\mathcal{N},
\end{equation}
which has the form of \eqref{smoothfeedback}. Stability and convergence properties of perturbed systems of the form \eqref{perturbed_closed} have been extensively studied in the control's literature \cite{sontag_robustness_noise}. Indeed, as discussed in \cite{sontag_robustness_noise}, \cite{sanfelice2006}, and \cite{poveda2021}, for the obstacle avoidance problem the disturbance $e$ can have a dramatic effect on the trajectories of the vehicle. To illustrate this fact, consider Figure \ref{fig:my_label}, where a vehicle, denoted with a white square, aims to converge to the target, denoted with a red circle while avoiding the obstacle $\mathcal{N}$ denoted with a white circle. Note that, to arrive at the target, the vehicle must choose a trajectory that goes above the obstacle or below the obstacle. Let $\mathcal{K}_1$ denote the set of initial conditions for which the closed-loop system \eqref{perturbed_closed} converges to the region $\mathcal{P}$ from above, and let $\mathcal{K}_2$ denote the initial conditions for which the closed-loop system \eqref{perturbed_closed} converges to the region $\mathcal{P}$ from below. It then follows that there must exist a set $\mathcal{K}$ where the vehicle must make a binary decision. Mathematically, for the obstacle avoidance problem, this behavior is captured by the following assumption; see also \cite{sanfelice2006,poveda2021}:

\vspace{0.05cm}
\begin{assumption} \label{assumption_robustness1}
There exists $T>0$ such that for each $\rho>0$ and each $\tilde{p}_0\in\mathcal{K}$, where $\mathcal{K}:=\overline{\mathcal{K}_1}\cap\overline{\mathcal{K}_2}$, there exist points $\tilde{p}_1(0), \tilde{p}_2(0)\in\{\tilde{p}_0\}+\rho\mathbb{B}$, for which there exist solutions $\tilde{p}_1$ and $\tilde{p}_2$ of \eqref{perturbed_closed} with $e=0$, satisfying $\tilde{p}_1(t)\in\mathcal{K}_1\backslash\mathcal{K}$ and $\tilde{p}_2(t)\in\mathcal{K}_2\backslash\mathcal{K}$ for all $t\in [0,T]$.  \begin{small}\QEDB\end{small}
\end{assumption}
\vspace{0.05cm}

Under Assumption \ref{assumption_robustness1}, the next proposition establishes zero margins of robustness against small adversarial perturbations $t\mapsto e(t)$ in the closed-loop system \eqref{perturbed_closed}. The proposition follows by \cite[Thm. 6.5]{SanfeliceDissertation} or \cite[Prop. 1]{poveda2021}:
\begin{proposition}\label{proposition_robustness}
Suppose that Assumption \ref{assumption_robustness1} holds. Then for each $\varepsilon$,$\rho'$,$\rho''>0$, and every $\tilde{p}_0\in \mathcal{K}+\varepsilon\mathbb{B}$ such that $\tilde{p}_0+\rho'\mathbb{B}\subset\mathbb{R}^2\backslash\mathcal{N}$ and $\tilde{p}_0+\rho''\mathbb{B}\subset\left(\mathcal{K}_1\cup\mathcal{K}_2\right)$  there exist a piecewise constant function $e:\text{dom}(e)\to\varepsilon\mathbb{B}$ and a (Carath\'{e}odory) solution $\tilde{p}:\text{dom}(\tilde{p})\to\mathbb{R}^2\backslash\mathcal{N}$ to \eqref{perturbed_closed} such that $\tilde{p}(t)\in(\mathcal{K}+\varepsilon\mathbb{B})\cap\left(\mathcal{K}_1\cup\mathcal{K}_2\right)\cap\left(\tilde{p}_0+\rho'\mathbb{B}\right)$, for all $t\in[0,T')$ for some $T'\in(T^*,\infty]$, where $\text{dom}~\tilde{p}=\text{dom}~\tilde{e}$, $T^*=\min\{\rho',\rho''\}m^{-1}$, and $m=\sup\{1+|k\nabla J(\eta)|:\eta\in p_0+\max\{\rho',\rho''\}\mathbb{B}\}$. If $T'$ is finite, then $\text{lim}_{t\to T'}\tilde{p}(t)\notin\left(\mathcal{K}_1\cup\mathcal{K}_2\right)\cup (\tilde{p}(0)+\rho'\mathbb{B})$. \QEDB
\end{proposition}

\vspace{0.1cm}
The result of Proposition \ref{proposition_robustness} has important implications for vision-based controllers based on perception maps, operating under topological obstructions such as obstacles. Namely, it establishes the existence of a set of points $\mathcal{K}\subset\mathbb{R}^2$ where \emph{arbitrarily small} approximations $e$ on the learned perception map $\ell$ can have a dramatic effect on the stability properties of the controller. Given that, in general, the error in the perception map $\ell$ can only be guaranteed to be bounded (see Lemma \ref{lemmaperceptionerrorbounded}), Proposition \ref{proposition_robustness} establishes that no robust controller based on smooth vector fields (e.g., based on navigation functions) exists for the solution of obstacle avoidance problems with inexact perception maps. Indeed, for navigation functions that combine attractive fields and repulsive fields, the set $\mathcal{K}$ will contain the spurious critical points of the navigation function $\phi$, which includes any saddle-point\footnote{The existence of such saddle points in navigation functions with attractive and repulsive fields was established in \cite{koditschek1990robot}.}. In this case, it is even possible to design adversarial disturbances $t\mapsto e(t)$ in \eqref{perturbed_closed} able to stabilize a spurious equilibrium \cite[Ex. 1]{poveda2021}.

\section{Robust Vision-Based Hybrid Control}
\label{sec_controller}
To synthesize a hybrid controller that overcomes the limitations of smooth feedback laws, we first characterize a class of admissible obstacles. We recall that $p_T\in\mathbb{R}^2$ denotes the target point of the robot.
\begin{assumption}\label{assumption_obstacle}
There exists $\rho \in \mathbb{R}_{>0}$ and $\varepsilon \in \mathbb{R}_{>0}$ such that the obstacle $\mathcal{N} \subset \mathbb{R}^2$ satisfies $\mathcal{N} \subset p_0 + \rho \mathbb{B}$ and $(p_0 + 2 \rho \sqrt{2} \mathbb{B}) \cap (\{p_T\} + \varepsilon \mathbb{B}) = \emptyset$, where $p_0 = [x_0, y_0]^T \in \mathbb{R}^2$. \hfill $\square$ 
\end{assumption}

\vspace{0.1cm}
In words, Assumption \ref{assumption_obstacle} considers obstacles that are contained in spheres located sufficiently far away from the target point.  Next, to achieve robust obstacle avoidance, we will design a switched \emph{perception-based} controller that implements different potential fields in different sub-regions of the operational space of the vehicle. By using this switching approach, we will be able to rule out the emergence of problematic sets $\mathcal{K}$ such as the one shown in Figure \ref{fig:my_label}. We note that our approach differs from existing works on hybrid control \cite{sanfelice2006,poveda2021,AvoidanceBisoffi} due to the use of perception maps employed by the vehicles to estimate their positions in real-time. However, we also stress that our methodology can be naturally extended to other hybrid controllers that are well-posed in the sense of \cite[Ch.7]{Goebel:12}.

\subsection{Synthesis of the Controller}
To design the covering of the operational space, for each $p_0 \in \mathbb{R}^2$ and $\rho > 0$, define the set $\mathcal{B}_{p_0, \rho}:= \{p \in \mathbb{R}^2 : ||p - p_0|| \leq 2 \rho \sqrt{2} \}$, which satisfies $\{p_0\} + \rho \mathbb{B} \subset \mathcal{B}_{p_0, \rho} \subset \{p_0\} + 2 \rho \sqrt{2} \mathbb{B}$. As in the standard state-based hybrid control \cite{sanfelice2006,poveda2021}, we define the sets:
\begin{align*}
    L_{1a} &:= \{p\in \mathbb{R}^2 : y < -x + y_0 + x_0 - 2\rho \sqrt{2} \},\\L_{1b} &:= \{p \in \mathbb{R}^2 : y < x + y_0 + x_0 + 2\rho \sqrt{2} \},\\
    L_{2a} &:= \{p \in \mathbb{R}^2 : y > x + y_0 + x_0 - 2\rho \sqrt{2} \},\\L_{2b} &:= \{p \in \mathbb{R}^2 : y > -x + y_0 + x_0 + 2\rho \sqrt{2} \},
\end{align*}
as well as the unions $\mathcal{O}_1:= L_{1a} \cup L_{1b}$, $\mathcal{O}_2 := L_{2a} \cup L_{2b}$, and $\mathcal{O}:= \mathcal{O}_1 \cup \mathcal{O}_2$. In this way, $\mathcal{O}=\mathbb{R}^2\backslash \mathcal{B}_{p_0, \rho}$ and $\mathcal{N}\cap\mathcal{O}=\{\emptyset\}$.  For each of the sets $\mathcal{O}_1$ and $\mathcal{O}_2$, we will design suitable potential functions $V_q$, $q\in\{1,2\}$, that can be used in a gradient-based controller of the form \eqref{closedloop1}. The controller will then switch between these two potential functions depending on its current location $p$ generated by a perception map $\ell$. Specifically, the potential functions are defined as
\begin{equation}\label{localization_function1}
V_{q}(p) :=  \begin{cases} 
      \phi_{q}(p)-\phi(p)~~~\forall~p\in\mathcal{O}_q \\
      ~~~~~~~\infty~~~~~~~~~\forall~p\notin\mathcal{O}_q,
   \end{cases}
\end{equation}
where $\phi$ and $\phi_q$ satisfy the next assumption.
\begin{assumption}\label{assumption_localization_functions}
The functions $\{V_q\}_{q\in\{1,2\}}$ satisfy the following: \textbf{(a)}  For each $q\in\{1,2\}$ there exist functions $\alpha_{1,q},\alpha_{2,q}\in\mathcal{K}_{\infty}$, and proper indicators\footnote{For a compact set $\mathcal{A}$ contained in an open set $\mathcal{U}$, a continuous function $\tilde{\omega}:\mathcal{U}\to\mathbb{R}_{\geq0}$ is a proper indicator of $\mathcal{A}$ on $\mathcal{U}$ if $\tilde{\omega}(z)=0$ if and only if $z\in\mathcal{A}$, and $\tilde{\omega}(z_i)\to\infty$ when $i\to\infty$ if either $|z_i|\to\infty$, or the sequence $\{z_i\}_{i=1}^{\infty}$ approaches the boundary of $\mathcal{U}$.} $\tilde{\omega}_q$ of $\{p_T\}$ on $\mathcal{O}_q$, such that $\alpha_{1,q}(\tilde{\omega}_q(p))\leq V_q(p)\leq\alpha_{2,q}(\tilde{\omega}_q(p))$,$\forall~p\in \mathcal{O}_q$; \textbf{(b)} For each $q\in\{1,2\}$, we have $\{p^*\in\mathcal{O}_q:\nabla V_q(p^*)=0\}={p_T}$; \textbf{(c)} For each $q\in\{1,2\}$, the function $V_q(\cdot)$ is continuously differentiable in $\mathcal{O}_q$. \QEDB
\end{assumption}
\begin{figure}
\begin{center}
\includegraphics[width=0.9\linewidth]{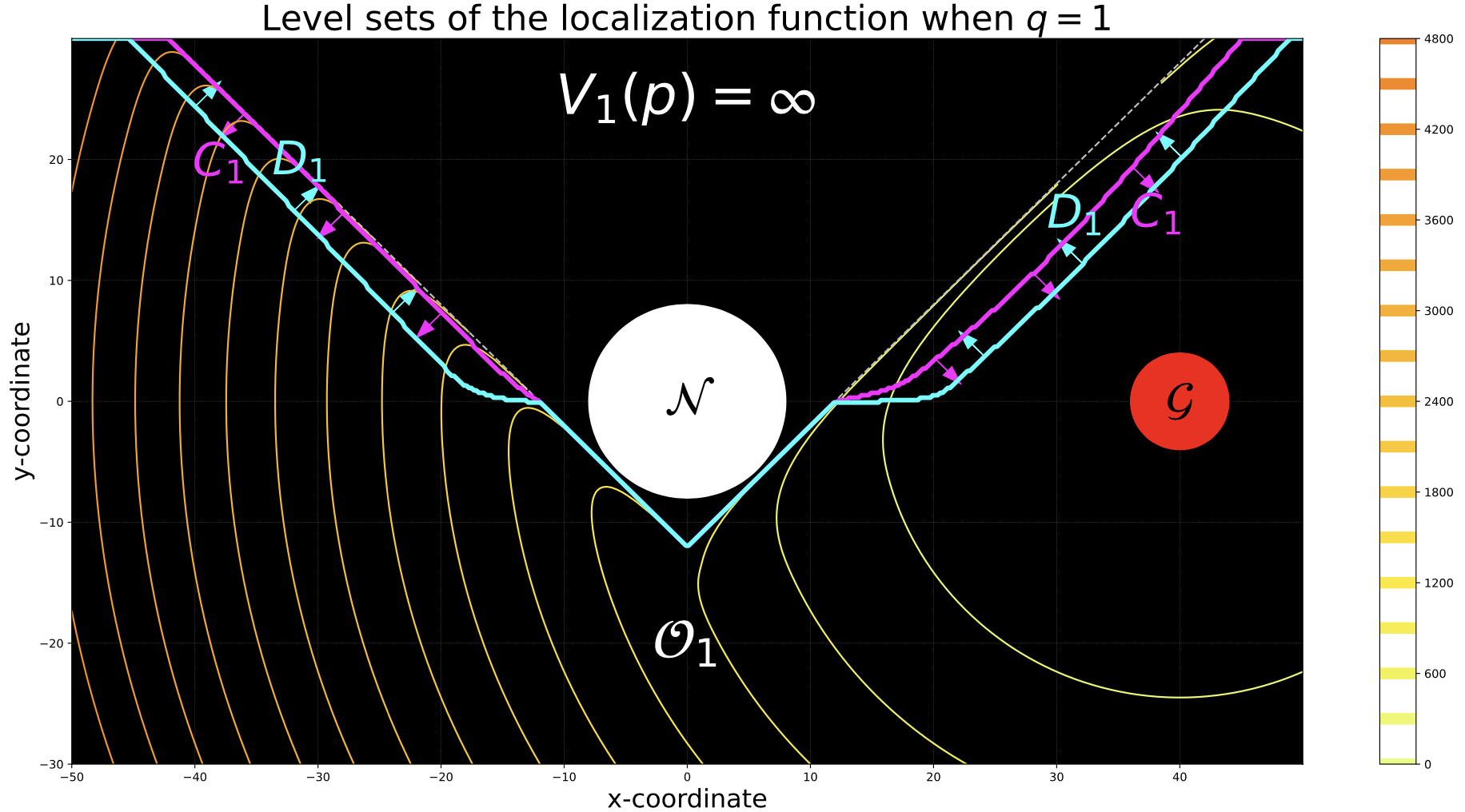}\\
\includegraphics[width=0.9\linewidth]{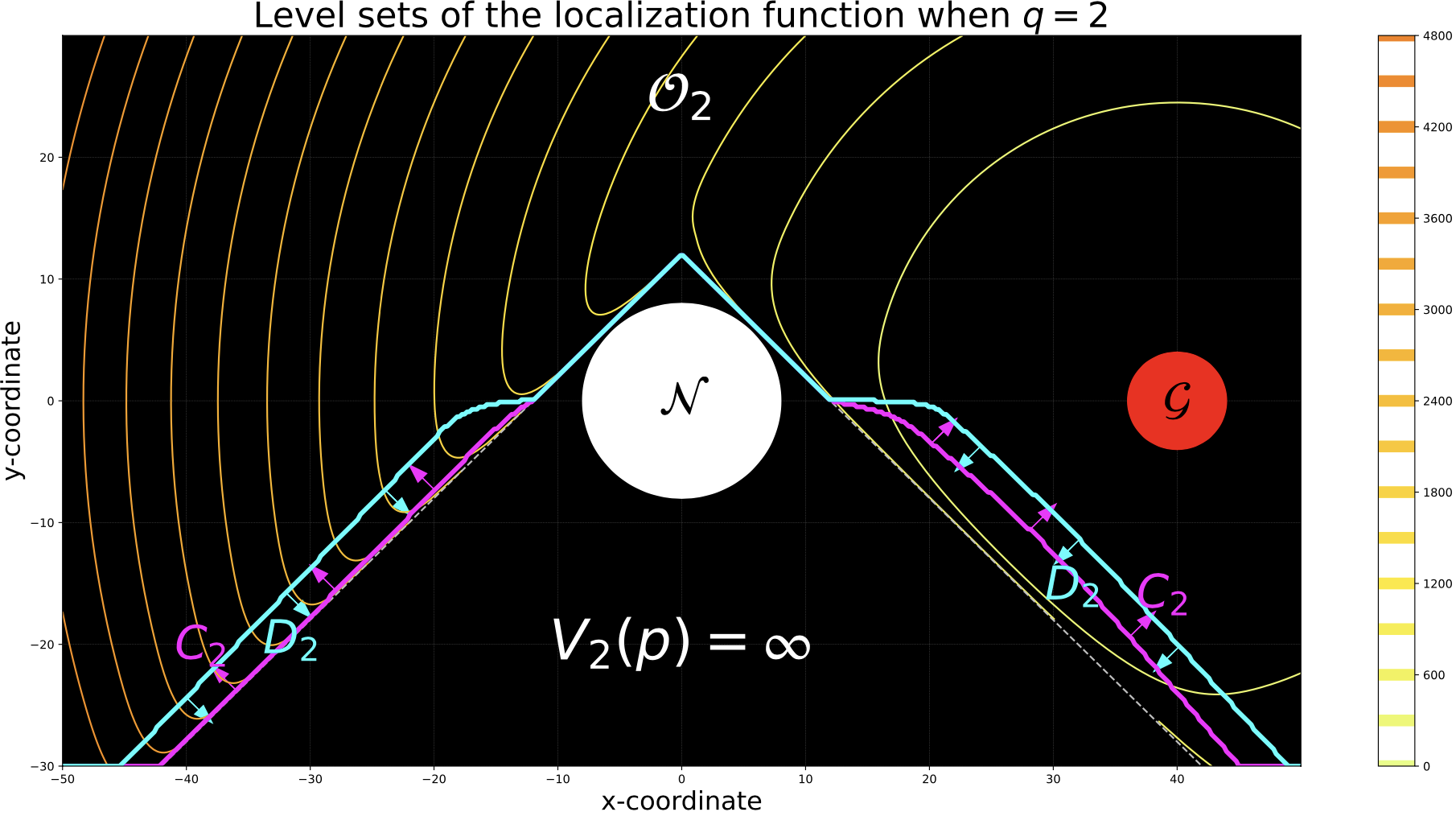}
\end{center}
   \captionof{figure}{\small{Covering of the operational space of the vehicle. Top: $q=1$. Bottom: $q=2$.}}
  \label{fig:coverings}
  \vspace{-0.4cm}
\end{figure}
\begin{remark}\label{remarkuseful}
A shown in \cite{poveda2021}, the conditions of Assumption \ref{assumption_localization_functions} can be readily satisfied using different classes of functions $\phi$ and $\phi_q$. For example, they hold when $\phi$ is given by $\phi=-(x-x_T)^2-(y-y_T)^2$, and $\phi_q$ is given by $\phi_{q}(p):=B(\tilde{d}_q(p))$, where $\tilde{d}_q(p):=|p|_{\mathbb{R}^2\backslash\mathcal{O}_{q}}^2$, and $B(s):=(s-\rho)^2\log\left(\frac{1}{s}\right)$, if $s\in[0,\rho]$, and $B(s):=0$, if $s>\rho$, with $\rho\in(0,1]$ being a tunable parameter selected sufficiently small. Figure \ref{fig:coverings} shows the geometric structure of both sets $\mathcal{O}_1$ (top plot), and $\mathcal{O}_2$ (bottom plot). The level sets of the functions $V_q$ are also shown in Figure \ref{fig:coverings}. Note that in each of the sets $\mathcal{O}_q$ the potential function $V_q$ has a unique critical point located at the position of the target.
\end{remark}
\begin{figure}
\centering
\includegraphics[width=0.8\linewidth]{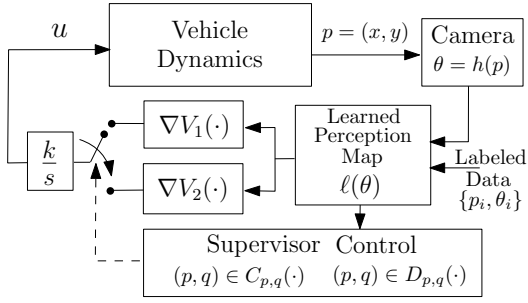}
   \captionof{figure}{\small{Scheme of the closed-loop system.}}
  \label{fig:scheme}
  \vspace{-0.5cm}
\end{figure}
\subsection{Main Results: Stability and Robustness}
Using the above construction, we can now formulate the complete perception-based hybrid control system. Let $\chi\in(1,\infty)$ and $\lambda\in(0,\chi-1)$ be tunable parameters. The closed-loop hybrid system has states $(p,q)\in\mathbb{R}^2\times Q$, where $Q=\{1,2\}$. The continuous-time dynamics are given by
\begin{equation}\label{gradient_flows1}
     \dot{p}=-k\nabla V_q(\ell(\theta)),~~~~~~\dot{q}=0,
\end{equation}
which are allowed to evolve in the set
\begin{equation}\label{flowset1}
    C_{p,q}:=\left\{(\ell(\theta),q)\in\overline{\mathcal{O}}\times Q:{V}_{q}(\ell(\theta))\leq \chi V_{3-q}(\ell(\theta))\right\}.
\end{equation}
The discrete-time dynamics are given by
\begin{equation}
     p^+=p,~~~q^+=3-q,
\end{equation}
which are allowed to evolve in the set 
\begin{equation}\label{jump_set1}
    D_{p,q} :=\left\{(\ell(\theta),q)\in \overline{\mathcal{O}}\times Q:{V}_{q}(\ell(\theta))\geq (\chi-\lambda)V_{3-q}(\ell(\theta))\right\}.
\end{equation}
Note that in \eqref{gradient_flows1}, \eqref{flowset1}, and \eqref{jump_set1}, the position of the vehicle is given by the perception map $\ell(\theta)$ rather than the state $p$. The term $(\chi - \lambda)$ in \eqref{jump_set1} guarantees that the intersection of the sets $C_{p, q}$ and $D_{p, q}$ is not empty. Thus, for initial conditions in $C_{p, q}\cap D_{p, q}$ solutions are not unique.  The set $C_{p, q}$  characterizes the points where the vehicle implements the controller \eqref{gradient_flows1} with constant state $q$. On the other hand, the set $D_{p, q}$ describes the points in the space where the vehicle toggles the logic state $q$ whenever it approaches the boundary of the respective set $\mathcal{O}_q$. In particular, note that since $\chi > 1$ and $\chi - \lambda > 1$, the robot toggles the potential field $V_q$ whenever its current value exceeds a threshold compared to the other potential fields $V_{q\prime}$. After each jump, the robot flows again using now the new potential function $V_{q\prime}$, until a new jump (if at all) is triggered. Note that this switching rule describes a hysteresis property in the feedback controller based on a supervisor mechanism. Figure \ref{fig:scheme} presents a schematic representation of the controller.

The following theorem is the main result of this paper. 
\vspace{0.1cm}
\begin{theorem}
Let $\delta>0$ and $K_0\subset C_{p,q}\cup D_{p,q}$, where $K_0$ is compact. Suppose that $p\mapsto F(p)$ is $L$-Lipschitz continuous, where $F$ is defined in Lemma \ref{lemmaperceptionerrorbounded}. Then, there exists a perception map $\ell$ and training data $\mathcal{T}=\{p_i,\theta_i\}_{i=1}^N$ such that every trajectory of the vehicle generated by the hybrid system \eqref{gradient_flows1}-\eqref{jump_set1}, with initial condition in $K_0$, is complete and converges to a $\delta$-neighborhood of the target point $p_T$ while avoiding the obstacle $\mathcal{N}$. \QEDB
\end{theorem}

\textsl{Proof:} Under Assumption 1, and using Lemma 1, for any compact set $K\subset\mathbb{R}^2$, and any pair $\varepsilon,L\in\mathbb{R}_{>0}$ there exists a perception map $\ell(\theta)$ satisfying the bound:
\begin{equation}
    |\ell(\theta)-p|\leq \varepsilon,~
\end{equation}
for all $p\in\mathcal{S}_{\varepsilon}^L$. It follows that $\ell(\theta)\in p+\varepsilon\mathbb{B},~~\forall~p\in\mathcal{S}_{\varepsilon}^L.$ Based on this observation, on compact sets, the solutions of \eqref{gradient_flows1}-\eqref{jump_set1} are also solutions of the inflated inclusion with state $z=(p,\theta)$:
\begin{subequations}\label{inflatedsystem}
\begin{align}
\dot{z}&=\left(\begin{array}{c}
\dot{p}\\
\dot{\theta}
\end{array}\right)\in F(z):=
\left(\begin{array}{c}
-k\nabla V_q(p+\varepsilon\mathbb{B})\\
0
\end{array}\right)\\
C_{p,q}&=\Big\{z\in\overline{\mathcal{O}}\times Q:{V}_{q}(p+\varepsilon\mathbb{B})\leq \chi V_{3-q}(p+\varepsilon\mathbb{B})\Big\}+\varepsilon\mathbb{B},\\ 
z^+&=\left(\begin{array}{c}
p^+\\
\theta^+
\end{array}\right)\in G(z):=
\left(\begin{array}{c}
p\\
3-q
\end{array}\right),\\
D_{p,q}&:=\Big\{z\in \overline{\mathcal{O}}\times Q:{V}_{q}(p+\varepsilon\mathbb{B})\geq\notag\\
&~~~~~~~~~~~~~~~~~~~(\chi-\lambda)V_{3-q}(\ell(\theta))\Big\}+\varepsilon\mathbb{B}.
\end{align}
\end{subequations}
In turn, every solution of \eqref{inflatedsystem} is also a solution of an inflated hybrid system, given by
\begin{subequations}\label{inflated_system}
\begin{align}
z&\in C_{\varepsilon},~~~~~\dot{z}\in F_{\varepsilon}(z),\\
z&\in D_{\varepsilon},~~~~~z^+\in G_{\varepsilon}(z),
\end{align}
\end{subequations}
where the data $(C_{\varepsilon},F_{\varepsilon},D_{\varepsilon},G_{\varepsilon})$ is defined as 
\begin{align*}
C_\varepsilon&:=\{z\in\mathbb{R}^n:(z+\varepsilon\mathbb{B})\cap C\neq\emptyset\},\\
F_\varepsilon(z)&:=\overline{\text{co}}~F((z+\varepsilon\mathbb{B})\cap C)+\varepsilon\mathbb{B}\\
D_\varepsilon&:=\{z\in\mathbb{R}^n:(z+\varepsilon\mathbb{B})\cap D\neq\emptyset\},\\
G_\varepsilon(z)&:=\{v\in\mathbb{R}^n:v\in g+\varepsilon\mathbb{B},g\in G((z+\varepsilon\mathbb{B})\cap D)\}.
\end{align*}
Based on this observation, in order to establish a stability property for the closed-loop system \eqref{gradient_flows1}-\eqref{jump_set1}, it suffices to establish a stability result for the inflated system \eqref{inflated_system}. 
The following Lemmas will be instrumental for our results.
\begin{lemma}
The closed-loop hybrid system \eqref{inflatedsystem} with $\varepsilon=0$ satisfies the Basic Conditions. \QEDB
\end{lemma}

\textsl{Proof:} Follows directly by \cite[Thm. 6.8]{Goebel:12}. \hfill $\blacksquare$

\begin{lemma}
Consider the HDS \eqref{inflatedsystem} with $\varepsilon=0$. Then, under Assumption \ref{assumption_localization_functions}, the set $\{p_T\}\times Q$ is asymptotically stable with basin of attraction given by $\mathcal{O}\times Q$.
\end{lemma}

\textsl{Proof}: The proof follows the same ideas of \cite[Sec. 6]{SanfeliceDissertation} and \cite[Thm. 1]{poveda2021}. Using Assumption 1, let us define $\tilde{\omega}(z):=\min_{q\in Q~s.t.~p\in\mathcal{O}_{q}}\tilde{\omega}_{q}(p)$ for each $z\in\mathcal{O}$. We obtain that $\tilde{\omega}$ is a proper indicator of $p_T$ on $\mathcal{O}$. Let
\begin{subequations}
\begin{align}
\alpha_1(s):=\min_{q\in Q}\alpha_{1,q}(s),              ~~~~\alpha_2(s):=\max_{q\in Q}\alpha_{2,q}(s),
\end{align}
\end{subequations}
Using Assumption \ref{assumption_localization_functions}-(a), the function $V_q$ satisfies 
\begin{equation}\label{upper_bound_L1}
\alpha_1(\tilde{\omega}(p))\leq V_{q}(p)\leq \alpha_2(\tilde{\omega}(p)),~~\forall~p\in\mathcal{O}.
\end{equation} 
During flows of the hybrid system, the time-derivative of $V_q$ is given by: 
\begin{equation}\label{inequality_Lyapunov}
\dot{V}_{q}(p)=-k|\nabla V_{q}(p)|^2<0,
\end{equation}
for all $(p,q)\in C_{p,q}\cap(\mathcal{O}\backslash\{p_T\})\times\{q\}$. Inequality \eqref{inequality_Lyapunov} implies that, for each $q\in\{1,2\}$, the function $V_q(z)$ decreases outside the target point $p_T$. 
\begin{figure*}
\begin{center}
\includegraphics[width=0.99\linewidth]{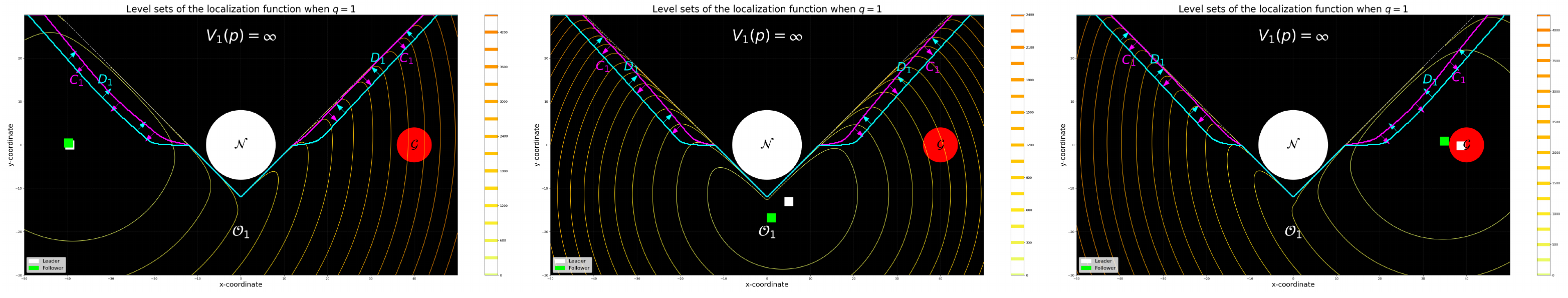}
\end{center}
   \captionof{figure}{\small{Time-varying level sets of a moving target converging to the static target $x_T$. In this simulation, a follower agent (green) tracks the  leader (white). The left, center and right columns show the level sets at the times 1, 90 and 180, respectively.}}
\label{fig:follower_level_sets14}
\vspace{-0.4cm}
\end{figure*}
On the other hand, jumps in the closed-loop system are allowed only when $V_{q}$ gets larger or equal than $(\chi-\lambda)V_{3-q}$. Since, by construction $(\chi-\lambda)>1$, it follows that during jumps $V_q$ satisfies:
\begin{equation*}
V_{q^{+}}(p^{+})\leq \frac{1}{\chi-\lambda} V_{q}(p),~~\forall~(p,q)\in D_{p,q}.
\end{equation*}
Therefore, the Lyapunov function decreases during jumps. The hysteresis mechanism rules out Zeno behavior, and the decrease of the Lyapunov function during flows and jumps implies that, for any complete solution of the system, the position $p$ converges to $p_T$, uniformly on compact sets in the basin of attraction \cite[Prop.7.5]{Goebel:12}. Completeness of solutions follows because: a) the system has no finite escape times; b) solutions cannot stop due to flows leaving the flow set; c) solutions cannot stop due to jumps leaving the union of the flow and jump set. This establishes the stability result. Obstacle avoidance follows by $\varepsilon$-closeness of solutions between the perturbed dynamics \eqref{inflated_system} and the nominal dynamics corresponding to \eqref{inflatedsystem} with $\varepsilon=0$. \hfill \QEDB

With Lemma 3 at hand, Theorem 1 follows now by a direct application \cite[Thm. 7.21]{Goebel:12}. \hfill$\blacksquare$ 

To the best knowledge of the authors, Theorem 1 is the first result in the literature that integrates perception-based maps and hybrid controllers with stability and convergence guarantees. In fact, the previous arguments can be trivially extended to guarantee robustness with respect to additional external disturbances, including small measurement noise, sporadic camera failures, and slowly moving targets (provided they remain in a compact set), a setting that emerges in leader-follower systems where the follower tracks the position of the leader. Figure \ref{fig:follower_level_sets14} illustrates this scenario by showing the level sets of $V_q$ at three different instants of time. Here, the white triangle denotes the position of the leader, which acts as a target for the follower, denoted with the green triangle.
\section{Numerical Experiments}
\label{sec_conclusions}
\begin{figure*}
\begin{center}
\includegraphics[width=0.75\linewidth]{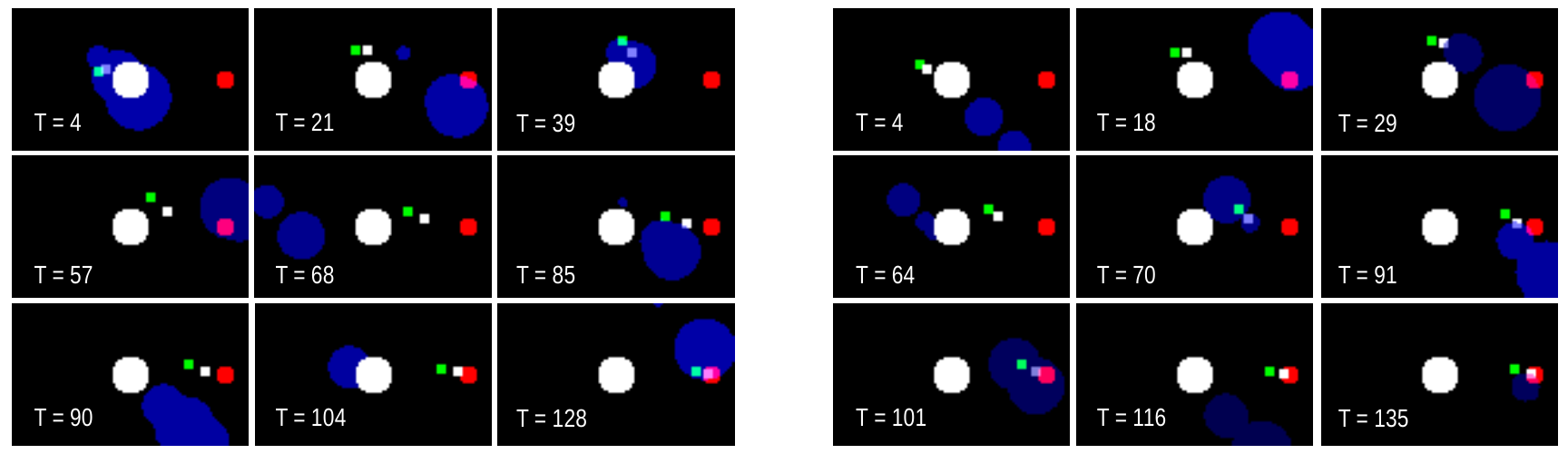}
\end{center}
   \caption{\small{Snapshots of the trajectories of the robots implementing the hybrid controller, but using different perception maps. In the left figure, the perception map was trained on images \textit{without} occlusions (shown in blue color). In the right plot, the perception map was trained using images \textit{with} occlusion. In both cases, the hybrid controller is successful.}
   }
\label{fig:results_occlusion}
\vspace{-0.4cm}
\end{figure*}

\begin{figure}[ht!]  
\begin{center}
\includegraphics[width=0.99\linewidth]{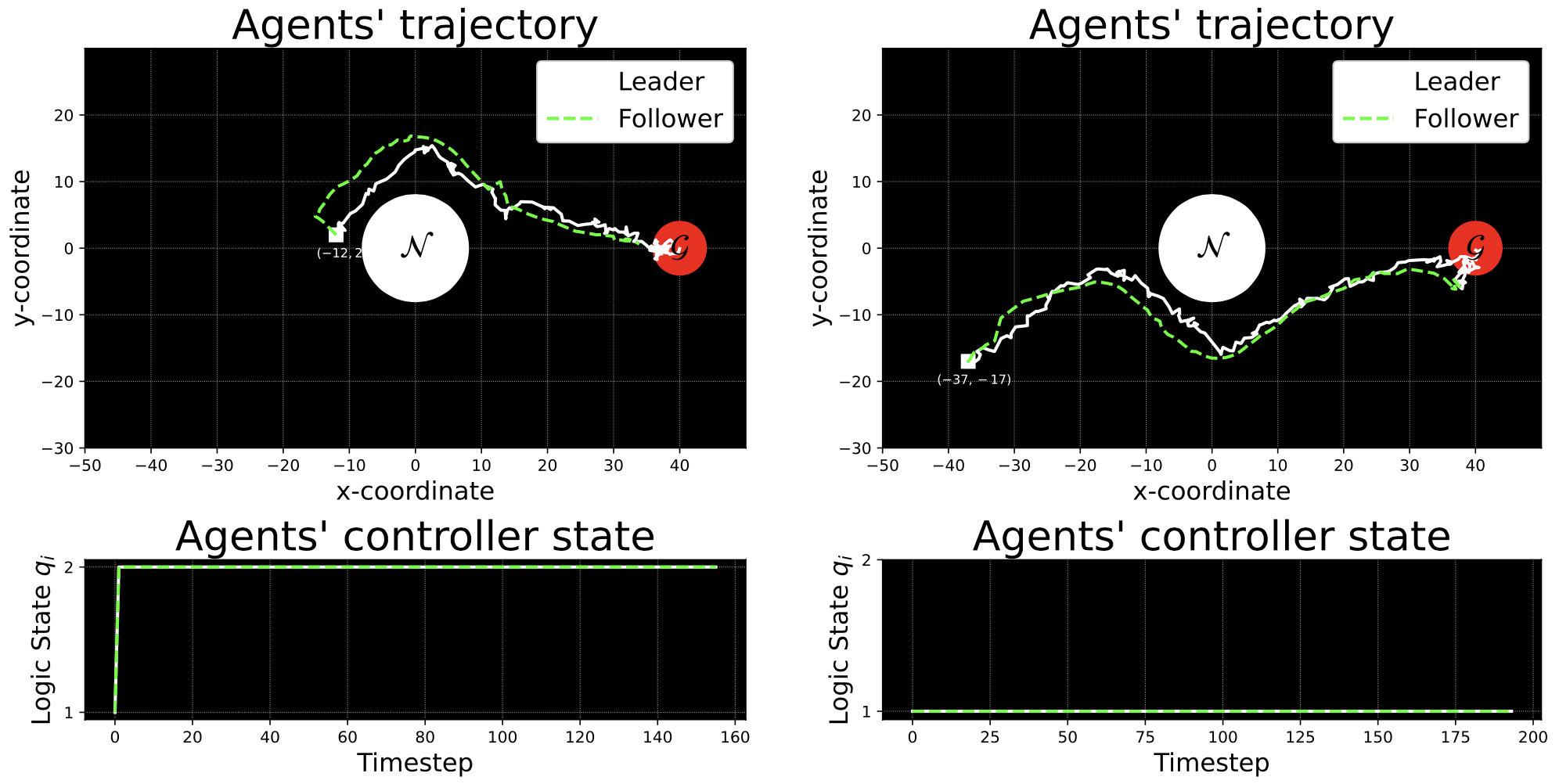}
\end{center}
   \caption{\small{Performance of the controllers when noise ($\epsilon \sim \mathcal{N}(0, 0.5)$) is added to the states of the robot, and the sensor of the follower has a failure rate of 50\% (no photo is taken). The flow and jump sets for this controllers are defined with $\chi = 1.1$ and $\lambda = 0.09$.}}
\label{fig:robustness_camerafailure+reference_noise}
\vspace{-0.5cm}
\end{figure}

We test the perception-based hybrid controller by training a perception map using a convolutional neural network. The model's architecture consists of three sets of Conv-ReLU-MaxPool blocks, with a kernel size of $3 \times 3$ and $2 \times 2$, respectively. A dense layer, preceded by a Dropout layer with probability 0.5, takes the flattened output of the last Conv-ReLU-MaxPool block of layers and outputs a vector that imperfectly describes the state information. The model was trained using Keras. The training took place for 5 epochs, with batch size 128 and input images with shape $(60, 100, 3)$. The output is a $1 \times 2$ vector describing the predicted $(x, y)$-position of the agent. The optimizer was Adam with learning rate $0.001$. The loss function was the Mean Squared Error (MSE) between the predicted and real state of the agent. Mean Absolute Error (MAE) was also used as a validation metric. To test the controller, we consider a leader robot (denoted with a white square) aiming to converge to the static target $\mathcal{G}$ while avoiding the obstacle $\mathcal{N}$. We also consider a follower robot (denoted with a green square), which tracks the leader. Both employ the hybrid controller, and the follower robot uses the learned perception map to approximate the leader's state. Figure \ref{fig:robustness_camerafailure+reference_noise} shows the trajectories obtained in this scenario under two different initial conditions: (-12, 2) in the left plot, and (-37, -17) in the right plot. In the simulations, we added noise to state measurements and we also included sensor failures (e.g., the camera does not transmit data at each time $t$ with a certain probability). As observed, the hybrid controller provides suitable robustness properties. On the other hand, in Figure \ref{fig:results_occlusion} we tested the generalization capabilities of the perception map. Here, we considered images containing occlusions (simulating, e.g., clouds). The left plot shows the performance of the hybrid controller when using a perception map trained on images \textit{without} occlusions. It can be seen that the controller successfully handles the increment in prediction error due to inputs to the vision model being generated from a different process. The right plot of Figure \ref{fig:results_occlusion} considers a controller trained on data \textit{with} occlusions. 

\section{Conclusions}
\label{conclusions}
In this paper, we introduced a perception-based hybrid controller for the robust solution of obstacle avoidance problems that use vision-based sensors for the purpose of feedback. Unlike existing results in the literature, our controller incorporates a perception map learned by using supervised learning methods, which provides a suitable approximation of the position of the vehicle based on images generated by a camera. By leveraging the structural robustness properties of the hybrid controller, and the generalization capabilities of the perception map, we established obstacle avoidance and convergence to the target point. Future research will focus on the theoretical guarantees under multiple obstacles.

\bibliographystyle{IEEEtran}
\bibliography{references}

\end{document}